Preprint of Gupta, S. and Degbelo, A. (2022) 'An empirical analysis of AI contributions to sustainable cities (SDG11)', in Mazzi, F. and Floridi, L. (eds) *The Ethics of Artificial Intelligence for the Sustainable Development Goals.*

# An Empirical Analysis of AI Contributions to Sustainable Cities (SDG11)

Shivam Gupta[1*] and Auriol Degbelo[2]

[1*]Bonn Alliance for Sustainability Research, University of Bonn, Regina-Pacis-Weg 3, Bonn, 53113, NRW, Germany.

[2]Institute of Geoinformatics, University of Münster, Schlosspl. 2, Münster, 48149, NRW, Germany.

*Corresponding author E-mail(s): shivam.gupta@uni-bonn.de;

Contributing author: auriol.degbelo@uni-muenster.de;

**Abstract**

Artificial Intelligence (AI) presents opportunities to develop tools and techniques for addressing some of the major global challenges and deliver solutions with significant social and economic impacts. The application of AI has far-reaching implications for the 17 Sustainable Development Goals (SDGs) in general and sustainable urban development in particular. However, existing attempts to understand and use the opportunities offered by AI for SDG 11 have been explored sparsely, and the shortage of empirical evidence about the practical application of AI remains. In this chapter, we analyze the contribution of AI to support the progress of SDG 11 (Sustainable Cities and Communities). We address the knowledge gap by empirically analyzing the AI systems (N = 29) from the AI×SDG database and the Community Research and Development Information Service (CORDIS) database. Our analysis revealed that AI systems have indeed contributed to advancing sustainable cities in several ways (e.g., waste management, air quality monitoring, disaster response management, transportation management), but many projects are still working *for* citizens and not *with* them. This snapshot of AI's impact on SDG11 is inherently partial yet useful to advance our understanding as we move towards more mature systems and research on the impact of AI systems for the social good.

**Keywords:** Artificial Intelligence, Sustainable Cities, AI for SDGs; Environment, Citizen Participation, SDG 11.

# 1 Introduction

Artificial intelligence (AI) has the potential to mitigate several issues facing cities, such as road safety, waste management, air pollution, and disaster risk reduction (Gupta et al., 2021). Examples of recent AI systems for improved well-being in cities include a tool for semi-automatic digitization of sketch maps to support the inclusion of indigenous communities through the documentation of their land rights (Degbelo et al., 2021; Chipofya et al., 2020), a system for traffic monitoring based on Wireless Signals (Gupta et al., 2018), approaches for efficient waste management (Barns, 2019), air quality modelling (Gupta et al., 2018) and urban health monitoring systems (Allam and Jones, 2020). Nonetheless, a lack of systemically observed knowledge and multidisciplinary perspective exists with limited coherence about the characteristics of AI contributions to sustainable cities (Zheng et al., 2020). Furthermore, as Israilidis et al. (2021) argued, the current research landscape is mainly focused on technical issues, leaving behind social impacts, participation capabilities, and knowledge sharing aspects with multi-stakeholder and citizen-inclusive development. Thus, the implementations of AI remain poorly understood.

To address this gap, this chapter looks into AI systems that contribute to advancing sustainable cities in several ways serving the Sustainable Development Goal (SDG) 11 proposed by the United Nations (UN) within the 2030 Agenda. The question asked is: What are AI4SG contributions for more sustainable cities in the digital age? AI4SG is defined in line with Cowls et al. (2021a) as the development of AI systems that enable socially preferable or environmentally sustainable developments. We look into the nature of the contribution of AI systems to more sustainable cities (what solution is proposed, to whom, and where) and the SDG indicators covered (which indicators are covered, which are still underrepresented). The analysis also covers the six citizen-centric challenges for smarter cities brought forth in Degbelo et al. (2016): the engagement of citizens, the improvement of citizens' data literacy, the pairing of quantitative and qualitative data to unlock new insight about city phenomena, the development of open standards, the development of personal services, and the development of persuasive interfaces, which can be supportive of inclusive progress towards SDG 11.

## 2 Related work

Cities are complex structures, growing worldwide at a fast pace (Batty, 2009). Commuter movement, capital flow, resources, and commodities lead to the emergence of city regions (Axinte et al., 2019). Due to increasing population size, density, and location, cities are also prone to adverse effects such as soil, air, and water pollution and impacts of climate change, affecting surrounding rural areas. Prompt action is required in the form of new and innovative infrastructures and services for addressing the increasing demands coupled with environmental and climate change impacts (Solecki et al., 2018).

Urban areas are increasingly digitalized over the last few decades due to significant advancements in digital technologies (Ismagilova et al., 2019). Cities are considered as the drivers for change and innovation (Fitjar and Rodríguez-Pose, 2020). Several innovative approaches are being developed to gather detailed insights and opportunities for the planning and management of cities (Sharda et al., 2021; Rogers et al., 2020). Notions such as *Smart cities* touched upon several dimensions or application domains where technological infrastructure, system integration, and data analysis can help us optimize resources in cities (Ismagilova et al., 2019). At the same time, cities are also trying to reconfigure themselves for a sustainable future, with the aim to improve the quality of life for all citizens (Barlacchi et al., 2015; Bibri, 2021). The importance of cities is well recognized by the internationally agreed Agenda 2030 Sustainable Development and the Paris Agreement to reduce the impact of climate change (Aust, 2019). In fact, two-thirds of all Sustainable Development Goals (SDGs) can only be achieved in and with the help of cities (Acuto, 2016). Emphasizing the opportunities offered by digital technologies at a city scale can significantly contribute towards the progress of sustainable development in line with the 2030 Agenda.

## 2.1 AI for SDG 11

Artificial intelligence (AI) and machine learning approaches are emerging as critical components for a smart and sustainable future by optimizing the services and addressing several social, environmental, and economic aspects in the cities (Allam and Dhunny, 2019). Thus, they could support progress towards SDG 11 (i.e., 'make cities and human settlements inclusive, safe, resilient and sustainable'). AI is fostering further advancements in technologies such as the Internet of Things (IoT), blockchain, robotics, precision health, and quantum computing (Firouzi et al., 2021; Dinh and Thai, 2018; Rajan and Saffiotti, 2017; Tajunisa et al., 2021; Dai, 2019), helping in making sense of large quantities of data by utilizing the innovation ecosystems that majorly exist in cities (Rabah, 2018). AI is instrumental in advancing the digitization processes in several cities (Sougkakis et al., 2020; Villagra et al., 2020; Majumdar et al., 2021), transforming them into more inclusive and sustainable environments. Advancements in Earth Observation (EO) technologies empowered with Artificial Intelligence (AI) is supporting various aspects of cities (Kuffer et al., 2020, 2021). From land use and pollutants monitoring in cities to supporting efficient energy and resource consumption (Yatoo et al., 2020; Shahid et al., 2021; Șerban and Lytras, 2020), AI provides us with the opportunities to address complex social inequalities and environmental interrelationships. Therefore, AI could be considered a crucial tool for addressing a wide array of challenges for future sustainable cities. Given the complexity and challenges of rapid urbanization, exploring the wide range of potential solutions across several domains may be desirable, as evident from the work mentioned above.

The possibilities offered by AI can only be utilized to their full potential for SDGs if the ethical, social, and environmental values are uniformly met (Hilbert, 2016; Gupta et al., 2021). The targets within the SDGs are intertwined as a unified framework in the form of 17 goals, forming an 'indivisible whole' (Nilsson et al., 2016). The goals and the targets are interlinked and depend on each other; but the views on how they are linked are still evolving (Nilsson et al., 2016; Vinuesa et al., 2020). Also, the capacity for integrating and intersecting intelligence from diverse domains for AI applications is growing. AI applications have the potential to make a significant contribution when several complex aspects are well integrated into the system for more inclusive action (Allam and Dhunny, 2019). There also exists a significant gap between cities having not made sufficient progress in such digitization sphere, creating a social divide and increasing inequalities (Reddick et al., 2020; Chase, 2020). The introduction of AI also risks amplifying some social and ethical challenges such as unfair bias, discrimination, or opacity in decision-making (Galaz et al., 2021). AI systems also require large amounts of energy and cause greenhouse gas (GHG) emissions (Taddeo et al., 2021; van Wynsberghe, 2021). Thus, highlighting that the application of AI and associated technologies, if not used thoughtfully, could also hurt social and economic aspects along with impacts on climate, biodiversity, and ecosystems around the world (van Wynsberghe, 2021). Therefore, it is crucial to be careful of the application of AI to ensure that efforts to harness the advantages of this technology outweigh its associated negative impacts.

## 2.2 Citizen centric approach for SDG 11

The aim of SDG 11 includes encouraging the development of cities and communities in a more inclusive, safe, resilient, and sustainable manner by making urbanization more inclusive for stakeholders, reducing the adverse effects of natural disasters, furthering local to global policies for sustainable development. SDG 11 addresses the urban level with ten targets and 15 indicators developed by the (United Nations, 2015). Implementation pathways lack comprehensive understanding, as coordination is required in terms of efforts from various stakeholders, embracing flexible and adaptive processes to accommodate changing circumstances, and allocating resources to address uncertain future threats, especially in the context of resilience (Croese et al., 2020). Limited evidence exists about the integration of genuine sustainability when we are more techno-centric, suggesting a knowledge-based development to address the existing complexities (Yigitcanlar et al., 2019). AI could support the progress of SDG11 through new solutions that enhance the food, health, transport, water, and energy services to the population. However, to date, less attention has been paid to the involvement of citizens in the process (Martens, 2019), which has enormous potential to contribute towards the SDGs progress by localization (Li et al., 2018).

AI systems enabling citizen participation can enhance the action towards sustainability through the collection of timely, high-resolution data, which could enhance the knowledge base required for SDGs progress (Fritz et al., 2019). Social and cultural information dictates the context in which the AI is implemented. Citizen participation provides the public with the opportunity to support policy development, leading to trust-building, credibility, and ultimately inclusiveness in taking actions towards SDGs. SDGs require actions that can transform existing practices across sectors. Fraisl et al. (2020) demonstrate that citizen participation "could contribute" to 76 indicators (33%) of SDGs, coverage of 60% indicators of SDG 11. It is crucial to integrate citizen-centric pathways to balance technological, social, and environmental factors (Kirwan and Zhiyong, 2020). The experientially trained or traditional or local knowledge from citizens could be a valuable source for addressing concerns related to disaster (Munsaka and Dube, 2018), urban planning (Antweiler, 2019), and environmental monitoring along with climate change mitigation (Makondo and Thomas, 2018; Magni, 2017). Citizen participation could act as relevant agents of change to mobilizing civil society for targets and indicators concerning sustainable consumption (Micheletti et al., 2014), air quality monitoring (Gupta et al., 2018), disaster risk mitigation (Ferri et al., 2020), sustainable and inclusive urbanisation (Newman et al., 2020). Multi-stakeholder participation and citizen-centered knowledge hubs could be instrumental for sustainable cities (Saner et al., 2020).

## 2.3 Exiting Gaps

AI is not the sole solution for developing sustainable cities, but as illustrated above, efforts to use AI for sustainable cities are increasing rapidly. These could help address complex challenges faced by humanity in social, environmental, and economic aspects (Vinuesa et al., 2020). If utilized carefully, outcomes supportive of sustainable development can be harnessed at a grand scale. Therefore, it is essential to learn the impact of AI as a tool for global good in a more systematic manner. Understanding this impact requires an understanding of factors that determine the advantage of using AI considered in a particular context as a part of sociotechnical systems (Cowls et al., 2021b). The SDGs here may provide a useful framework.

Nevertheless, SDGs are sometimes considered ambitious and wide-ranging (Pekmezovic, 2019). This ambitious and wide-ranging nature also inspires and stimulates action for sustainable development (Walker et al., 2019). Several systematic approaches were undertaken in the recent past to gather evidence of the use of AI for SDGs worldwide, resulting in the generation of datasets and knowledge bases organized in different forms, presenting a distinct picture of the impact AI has on SDGs (Vinuesa et al., 2020; Tomašev et al., 2020; Cowls et al., 2021b; Palomares et al., 2021). However, it is imperative to note that these studies reflect on the impact of AI for SDGs at a high level and often include evidence from experimental closed systems. A deeper analysis is required to understand the role of different actors, practitioners, impacts, social implications, and contribution of AI to specific sub-goals of SDGs, which is scarcely discussed. Additionally, understanding discrepancies between the relevance of AI at the goal level and deeper conflicts amid the need for SDG targets and indicators is essential to realize how key stakeholders could carefully use AI for sustainable development.

Overall, flawed understanding of complexities in implementation of AI systems for SDGs, governance hurdles, lack of knowledge about the influence of AI on suitable targets and indicators, unclear role and responsibilities among stakeholders lead to uncoordinated exercises, thus limiting us from realizing the full potential of technological innovation for sustainable development. Additionally, civic awareness, citizen engagement, ownership, and citizen-centric approaches must be enhanced for inclusive action (Guan et al., 2019; Rubio-Mozos et al., 2019; Thinyane, 2018). The remainder of the chapter intends to inform discussions on both AI for SDG11 and for deeper citizen engagement through a systematic analysis of the contributions of past and ongoing projects.

## 3 Method

We critically analyzed existing projects on AI4SG and CORDIS database to synthesize progress and learn about current gaps. The data collection and analysis were done in four steps.

**Step 1: Data Collection AIxSDGs.** We have retrieved all projects from the Oxford Initiative on AIxSDGs, which are related to SDG11. Contrary to the AIxSDGs initiative, which did the mapping at the goal level, the mapping in this work was done at the indicator level.

**Step 2: Data Collection CORDIS.** We have retrieved all projects from the CORDIS database that are related to the theme of the paper. There are 12 possible contributions to search for in the database: 'Projects', 'Results Packs', 'Research*EU Magazines', 'Results in Brief', 'News', 'Events', 'Interviews', 'Report summaries', 'Project Deliverable', 'Project Publications' 'Exploitable Results', and 'Programmes'. We decided to focus on 'Project Deliverables', 'Project Publications' and 'Exploitable Results', since we are interested in concrete outcomes. Also, focusing on these types of contributions is consistent with the data obtained from AIxSG (Step1) because the projects obtained from Step 1 share the common feature that they have been successfully implemented on the ground for at least six months and have no negative impact measured. Besides, the CORDIS Web application enables the search of results by application domain and offers 11 application domains: 'Industrial Technologies', 'Fundamental Research', 'Transport and Mobility, 'Health', 'Society', 'Security', 'Climate Change and Environment', 'Energy', 'Space', 'Digital Economy', and 'Food and Natural Resources. The two authors went through the 10 targets of SDG 11 and mapped them to the 11 themes of the CORDIS platform. The results of the mapping were: 11.1 => (NA), 11.2 => (Transportation mobility), 11.3 => (Society), 11.4 => (NA), 11.5 => (Climate change and environment), 11.6 => (Climate change and environment), 11.7 => (Society), 11.a => (NA), 11.b => (NA), and 11.c => (NA).

As a result, we searched for the project deliverables, project publications and exploitation results related to the themes 'Transport and Mobility, 'Climate Change and Environment and 'Society'. 'Artificial Intelligence' or AI system can be defined in many ways as shown in (Samoili et al., 2020). Hence, the search for AI-related work in the CORDIS database was done using a variety of keywords. We have used two sources for these keywords: keywords pointing at the sub-domains of AI suggested by the Joint Research Centre (Samoili et al., 2020) and keywords from the AI Glossary by (Hutson, 2017). The search strings used were:

- JRC subdomains search string: "Knowledge representation" or "Automated reasoning" or "Common sense reasoning" or "Planning" or "Scheduling" or "Searching" or "Optimisation" or "Computer vision" or "Audio processing" or "Multi agent systems" or "Robotics" or "Automation" or "Connected vehicles" or "Automated vehicles" or "AI Services" or "AI Ethics" or "Philosophy AI".
- AI glossary search string: "Algorithm" or "Backpropagation" or "Black Box" or "Deep Learning" or "Expert System" or "Generative Adversarial Networks" or "Machine Learning" or "Natural Language Processing" or "Neural Network" or "Neuromorphic Chip" or "Perceptron" or "Reinforcement Learning" or "Strong AI" or "Supervised Learning" or "Tensorflow" or "Transfer Learning" or "Turing Test".

The search on October 3, 2021 returned 333 results.

**Step 3: Filtering.** The results obtained from the CORDIS database were filtered to keep only the projects that have developed AI systems. At this stage, some outcomes (N=320) from the previous step were excluded, and N=13 results were included in the final analysis. 16 projects were identified from the AIxSDG database. At the end of this step, 29 projects remained (see Table 1), which were included in the final analysis.

**Step 4: Coding.** For each project selected (step 1 and 3), we coded the nature of the contribution (what solution is proposed, to whom and where), the SDG indicators covered (the indicators to which the AI system proposed is relevant), and the citizen-centric challenges to which the AI system is relevant. The coding was done deductively and went through many iterations (i.e., the categories were defined a priory based on the existing scientific and grey literature, and we remained open to extending the original list during the coding if some categories were missed).

Autonomous Vehicle (i.e., self-driving cars, autonomous drones). As for the *beneficiary*, we used a relatively-coarse categorization based on who pays for the product or system: companies/businesses, government/public sector, and citizens. Prototypes developed during research projects, unless they have a dedicated citizen-focus, fell under the category of government/public sector. Deciding on the *cities where the solution was deployed* proved to be a challenge because of the varying level of granularities at which the projects were documented. At times, the location where the solution was deployed was not at all reported. At other times, the solution was deployed in several cities (again here, datasets on the exact locations where it has been deployed were not available or sparsely available). For this reason, we had to resort to some simple rules: 1) include the city when it is explicitly mentioned in the project description or some supplementary material (e.g., demo video) on the Web; 2) when the system has been deployed in many cities (e.g., the *Optibus* project)[1], the city of the headquarter is used as a location for the project; 3) research projects documented in the CORDIS database often did not report on the deployment sites or had used several sites for cross-validation as is typically the case for European projects. Consistent with the use of the headquarter of the companies above, we have used the headquarters' location of the coordinating institution of the project. The list of *SDG indicators* was taken from the UNDESA SDG Indicators Metadata repository (United Nations Department of Economic and Social Affairs (UNDESA), 2015; UNDESA Statistics Division, 2021). Finally, the definition of *citizen-centric challenges* was taken from (Degbelo et al., 2016): deep participation (i.e., working with citizens, not only for them), the data-literate citizenry (i.e., promotion of data literacy skills and the fostering of digital inclusion), pairing quantitative and qualitative data (i.e., the combination of quantitative data with volunteered geographic information by users, that is typically qualitative), open standards (i.e., data available as open data, along with the development or promotion of open standards for data collection, analysis, storage, and sharing), personal services (i.e., services adaptive to the abilities, expertise, and needs of individual users), and persuasive interface (i.e., interfaces that raise awareness about, stimulate or encourage change towards more sustainable behaviors). The results of the coding are presented next.

[1] https://www.aiforsdgs.org/all-projects/optibus.

## 4 Results

We now report on the outcomes of the coding process. The reporting presents some descriptive statistics about the geographic distribution of the projects examined, their key beneficiary, the type of system developed, the target and indicators for which they are relevant, and the citizen-centric challenges they connect to. Interpreting the data and providing some speculative implications is done in Section 5. To facilitate readability, the name of the project is left in CAPITALS when the original project acronym was provided in capitals. Else the name of the project is italicized.

### 4.1 Geographic distribution of AI projects

Figure 1 shows the geographic distribution of the different projects. A safe interpretation of this map (given the different possible interpretations of location, see the discussion in Section 3) is that it gives an idea about where past/ongoing AI-powered projects related to the SDG 11 have been *initiated*.

### 4.2 Key beneficiaries of AI projects

The majority of the projects (76%) in the datasets were targeted at the government or the public sector. Examples of this type of project include the *Prometea* project that led to substantial time savings in the Argentinian judicial system, projects that try to enhance the public transportation infrastructure through the use of autonomous cars (e.g., *Cybermove*, *FiveAI*), and several projects that attempt to address the problem of environmental monitoring from different angles (e.g., *ENVISNOW* for monitoring snowmelt, *CAMELS* for monitoring terrestrial carbon sink, and *SPHERE* for flood risk estimation). 24% of the projects targeted improvements for companies/businesses. Examples of these projects include those attempt to address the issue of efficient energy management in cities (e.g., *Ennet Eye* for building energy management, *AIxAI* for efficient transportation/energy resource distribution), and projects that attempt to improve waste treatment and management (e.g., the *IRBin*, *RUBSEE*). Overall, only a few projects (10%) can be said to address the needs of the civil society: *Qucit* has developed tools to facilitate the finding of parking spaces and bikes in cities; the *National Fine Dust Forecast Project* provided applications to inform citizens about the concentration of air pollutants, helping thereby better protect themselves against these pollutants; and the *Breeze* project strives to provide better information about air quality through its platform.

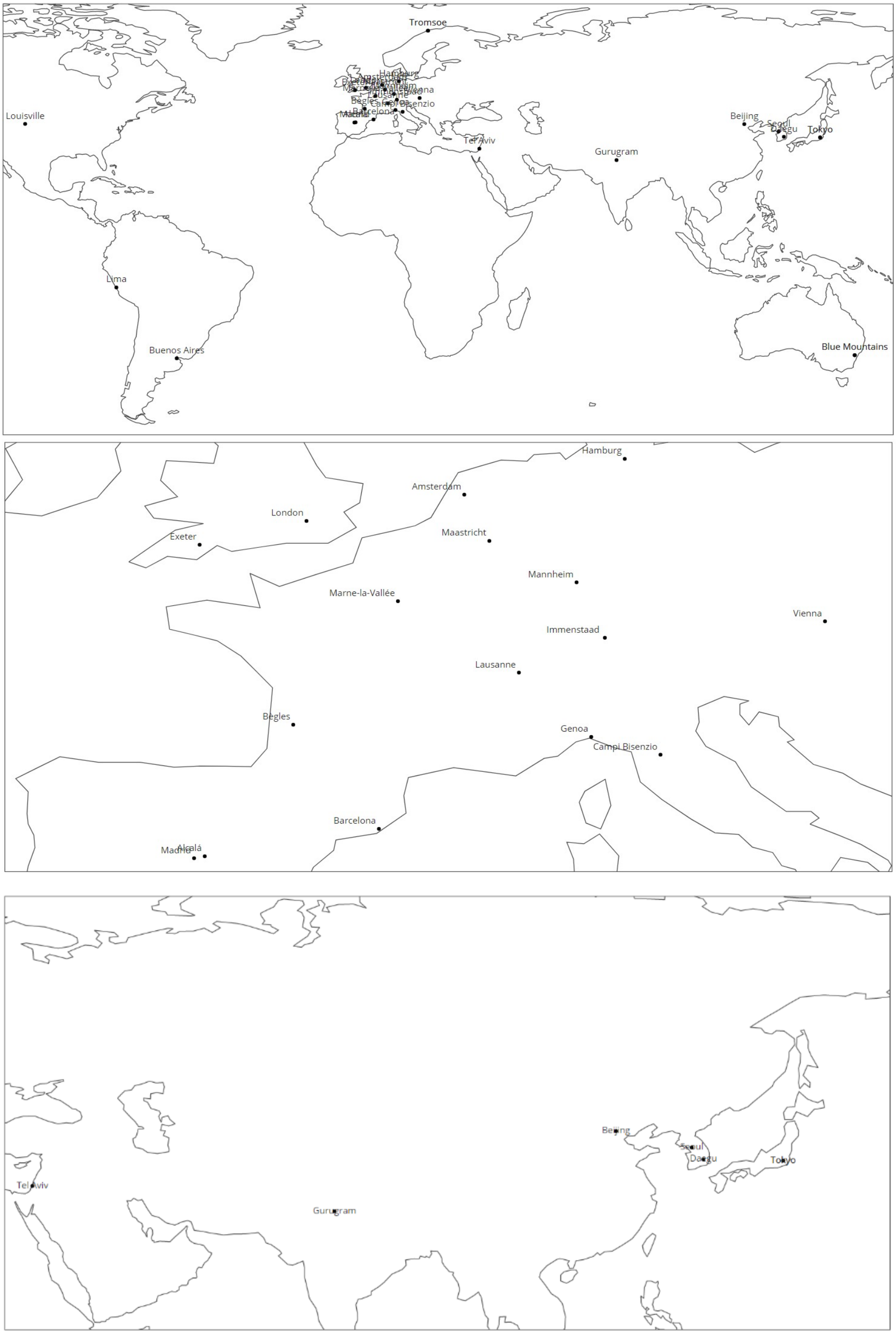

**Fig. 1** Geographic distribution of the places where projects in the datasets have been initiated (Top: overview; Middle: zoom on central European Countries; Bottom: zoom on Asian Countries).

**Table 1**: Overview of the projects and their contributions. Legend: G/PS (Government/Public Sector), C/B (Company/Business), AI4SDG (Data from the AI4SDG database), CORDIS-G (Data from the CORDIS database, obtained after the search using the Keywords from the AI Glossary, CORDIS-J (Data from the CORDIS database, obtained after the search using the Keywords from the JRC).

| Project name | Type of system | Key Beneficiary | Target | Social Impact | Dataset |
|---|---|---|---|---|---|
| IRBin | Robot | C/B | 11.6 | Efficient municipal waste management | AI4SDG |
| Prometea | Software Application | G/PS | 11.b | Resource efficiency (substantial time savings) in the judicial system | AI4SDG |
| Brightics AI | Software Application | C/B | 11.5 | Enhanced risk analysis (natural disasters, weather, social issues) | AI4SDG |
| National Fine Dust Forecast Project | Software Application, Analysis Model | G/PS, citizens | 11.6 | Improved citizens' protection against air pollutants | AI4SDG |
| Ennet Eye | Software Application | C/B | 11.b | Enhanced building energy management | AI4SDG |
| AIxAI | Software Application | C/B | 11.b | Efficient resource distribution (transportation, energy) | AI4SDG |
| UNIST Heatwave Research | Analysis Model | G/PS | 11.3 | Better informed Human settlement planning | AI4SDG |
| FiveAI | Autonomous Vehicle | G/PS, citizens | 11.2 | Enhanced public transportation infrastructure | AI4SDG |
| Optibus | Software Application | C/B | 11.2 | Optimized transit in cities | AI4SDG |
| Seneka | Robot | G/PS | 11.5 | Faster disaster rescue operations | AI4SDG |
| Breeze | Software Application | citizens, G/PS | 11.6 | Better informed air quality monitoring | AI4SDG |
| Qucit | Software Application | citizens, G/PS | 11.7 | Improved resource finding (parking spaces, bikes) | AI4SDG |
| RUBSEE | Robot | C/B | 11.6 | Improved waste treatment | AI4SDG |
| AMP Robotics | Robot | C/B | 11.6 | More efficient recycling (plastic, metals) | AI4SDG |
| DiDi Smart Transportation Brain | Software Application | G/PS | 11.2 | Enhanced transportation services | AI4SDG |
| Dynamic and Robust Wildfire Risk Prediction System | Analysis Model | G/PS | 11.5 | Enhanced risk analysis (wildfire) | AI4SDG |
| MEDACTION 4 | Analysis Model, Software Application | G/PS | 11.3 | Desertification management strategies | CORDIS-G |
| CLEOPATRA | Software Application | G/PS | 11.5 | Enhanced oil pollution monitoring | CORDIS-G |
| REVAMP | Software Application | G/PS | 11.5 | Better informed coastal disaster emergency management | CORDIS-G |
| DAYWATER | Software Application | G/PS | 11.5 | Improved urban storm water monitoring | CORDIS-G |
| ENVISNOW | Analysis Model, Algorithm | G/PS | 11.5 | Enhanced modelling of snowmelt | CORDIS-G |
| FLOODMAN | Analysis Model | G/PS | 11.5 | Improved monitoring of water bodies | CORDIS-G |
| Cybermove | Autonomous Vehicle | G/PS | 11.2 | Enhanced public transportation infrastructure | CORDIS-J |
| geoland | Analysis Model | G/PS | 11.3 | Geoinformation services for land monitoring | CORDIS-J |
| SITAR | Software Application | G/PS | 11.6 | Improved monitoring of toxic waste | CORDIS-J |
| CAMELS | Analysis Model | G/PS | 11.3 | Estimation of terrestrial carbon sink | CORDIS-J |
| MEGAFIRES | Analysis Model | G/PS | 11.5 | Improved risk estimation (wildfire) | CORDIS-J |
| ECOSIM | Analysis Model | G/PS | 11.6 | Improved air quality forecasting | CORDIS-J |
| SPHERE | Analysis Model, Software Application | G/PS | 11.5 | Enhanced flood risk estimation | CORDIS-J |

## 4.3 Types of systems

Robots (14%) are one type of AI contribution to more sustainable cities. They have been deployed to facilitate waste management (as done, for instance, in the *RUBSEE*, *AMP Robotics*, and *IRBin projects*) or to facilitate rescue operations during disaster management (e.g., the *Seneka* Project). Other contributions are made in the form of software applications, for instance, to facilitate data analysis through an (analytics) platform (see the *Brightics AI* project) or to speed up work in the judicial domain (e.g., the *Prometea* project). Software Application contributions were more frequent in the dataset (52%). Another type of contribution (38%) is in the form of Analysis Models (e.g., to predict heatwaves, see the *UNIST* Heatwave Research project). 2 projects (*Cybermove* and *FiveAI*) are concerned with self-driving cars, and one project (i.e., the *ENVISNOW* project) proposed an algorithm to retrieve snow depth by using artificial neural networks and multi-frequency radiometric data from satellites.

## 4.4 Targets served by AI projects

Target 11.5 appears more frequently (34%) in the dataset as a result of several projects dealing with disaster mitigation and management as a use case, e.g., the *Seneka* project mentioned above, the *Dynamic and Robust Wildfire Risk Prediction System* (predicting wildfire risk from weather data, see (Salehi et al., 2016)), the *MEGAFiReS* project (fire monitoring with remote sensing images), the *FLOODMAN* project (flood monitoring), and the *CLEOPATRA* project (oil and marine pollution). A share of projects (21%) is concerned with shaping more environmentally friendly cities (SDG Target 11.6) through improved waste management/treatment (e.g., *IRBin*, *AMP Robotics*, *SITAR*), or providing 'better' information regarding the quality of the air (e.g., the *National Fine Dust Forecast Project* or *Breeze*). Our sample had an equal share of projects dedicated to Target 11.2 (sustainable transport systems, 14%) and Target 11.3 (sustainable human settlement planning, 14%). Past/ongoing AI systems relevant to target 11.2 have been introduced (or are being explored) to optimize transit in cities (e.g., the *Optibus* project), expand the existing transportation infrastructure through the use of autonomous vehicles (e.g., *Cybermove*, *FiveAI*), or services to facilitate the management of the traffic flow (e.g., *DiDi Smart Transportation Brain*). The four projects relevant to target 11.3 in our sample contributed with management strategies for desertification (e.g., *MEDACTION 4*), tools to inform improved human settlement planning (e.g., the *UNIST Heatwave Research* project), and built tools/models that could be used for improved urban planning (e.g., *geoland* proposed the Observatory Spatial Planning to 'put[...] urban growth on the map', and *CAMELS* proposed models for the terrestrial carbon sinks). The remaining three projects contribute to more efficient resource management (i.e., Target 11.b for *Ennet Eye* and *AIxAI*) and the unlocking of new possibilities to access urban spaces (i.e., Target 11.7 for *Qucit*). The connection of the AI projects to the SDG11 targets is shown in Figure 2.

## 4.5 Indicators supported by AI projects

The relevance of the AI projects to the SDG11 indicators are shown in Figure 3. A glimpse at the figure shows that the data is more skewed towards Indicator 11.5.2 (Target 11.5) and has an almost equal share of items for the Indicators 11.6.1 and 11.6.2 (Target 11.6). More interesting here is that not all projects connected to a target could be assigned to an indicator. This is an issue that may point at the need to expand the list of indicators to cover (important) aspects of sustainable cities that are not covered currently. To inform future work along those lines, we report on why these projects fit a target but do not fit an indicator.

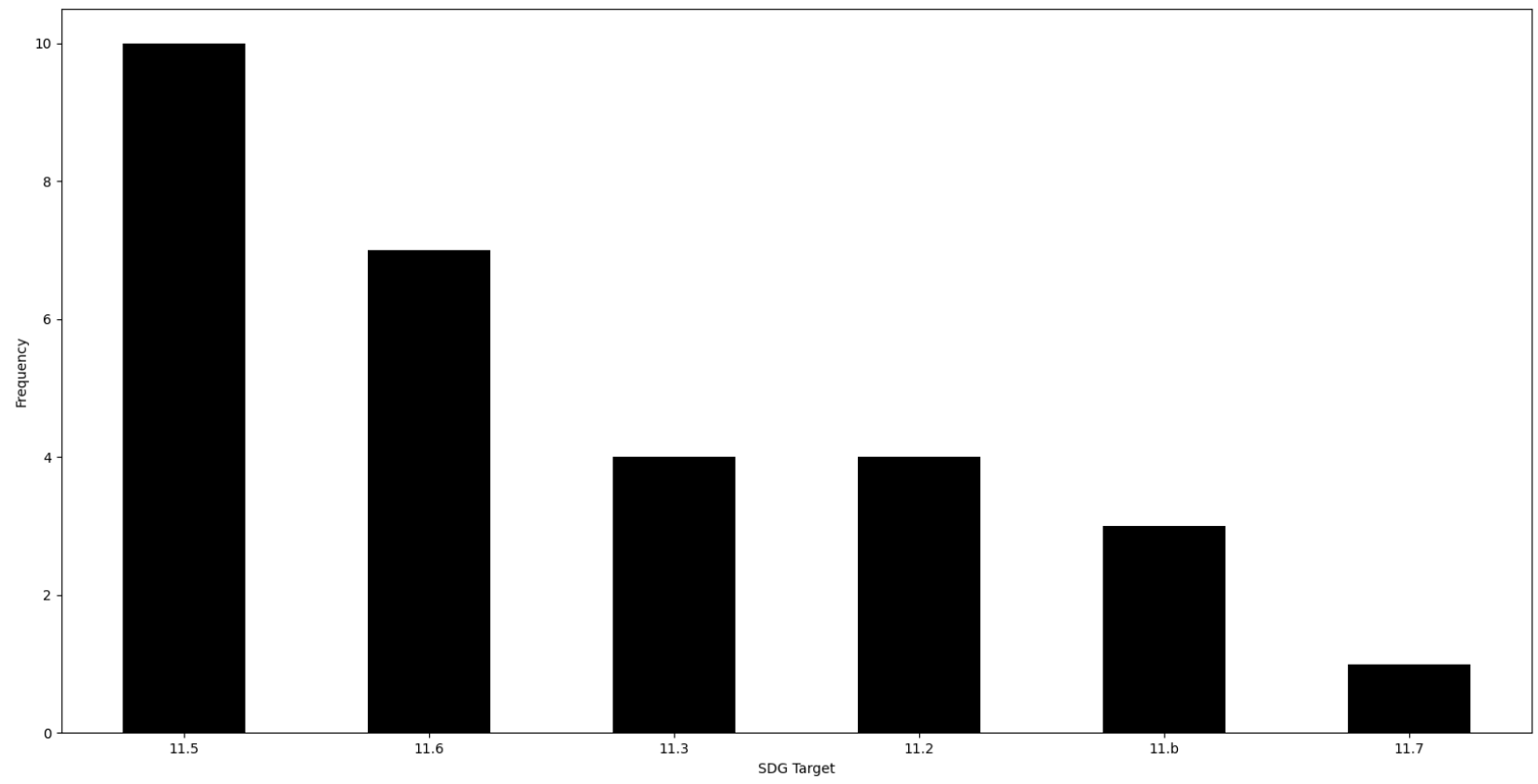

**Fig. 2** Projects in the dataset and their relevance to the SDG 11 targets.

- *Ennet Eye* (Target 11.b): the system proposed detects "problems such as the unnecessary use of electricity and presents the economic burden and possible solutions to the problems in order to improve energy efficiency"[2]. As such, it is a useful solution towards sustainable resource usage, but none of the two indicators 11.b.1 (number of countries that adopt national disaster risk reduction strategies) and 11.b.2 (proportion of local governments that adopt and implement local disaster risk reduction in line with national disaster risk reduction strategies) would have done justice to that aspect of sustainable electricity usage in the city. This would have fallen rather under SDG7. Target 7.3 reads: "By 2030, double the global rate of improvement in energy efficiency" and the Indicator 7.3.1 reads: "Energy intensity measured in terms of primary energy and GDP". A question this raises is whether or not some indicators directly relevant to energy efficiency are needed for monitoring progress on sustainable cities;

[2] https://www.aiforsdgs.org/all-projects/ennet-eye-powered-energylink.

- *AIxAI* (Target 11.b): The project allows "real-time area management for efficient resource distribution"[3]. Examples of 'resources' mentioned in the project description include air conditioning, operation of elevators and escalators, cleaning and personnel costs. The argument stated above regarding indicators related to the improved energy efficiency in cities applies;
- *Prometea* (Target 11.b): the project achieved substantial time-saving gains through the introduction of digitization/AI in the judicial system[4]. This is an indirect contribution to the mitigation of climate change. A non-digital process implies much being done *in-person* and several rounds of in-person travelling to get some information or provide information (e.g., a missing document for the process). A digital process reduces the cost of accessing and exchanging information and reduces the need for travelling/commuting to the benefit of the climate. There is currently no indicator covering the (positive/negative) impact of digitization on climate change (e.g., $CO_2$ emissions);

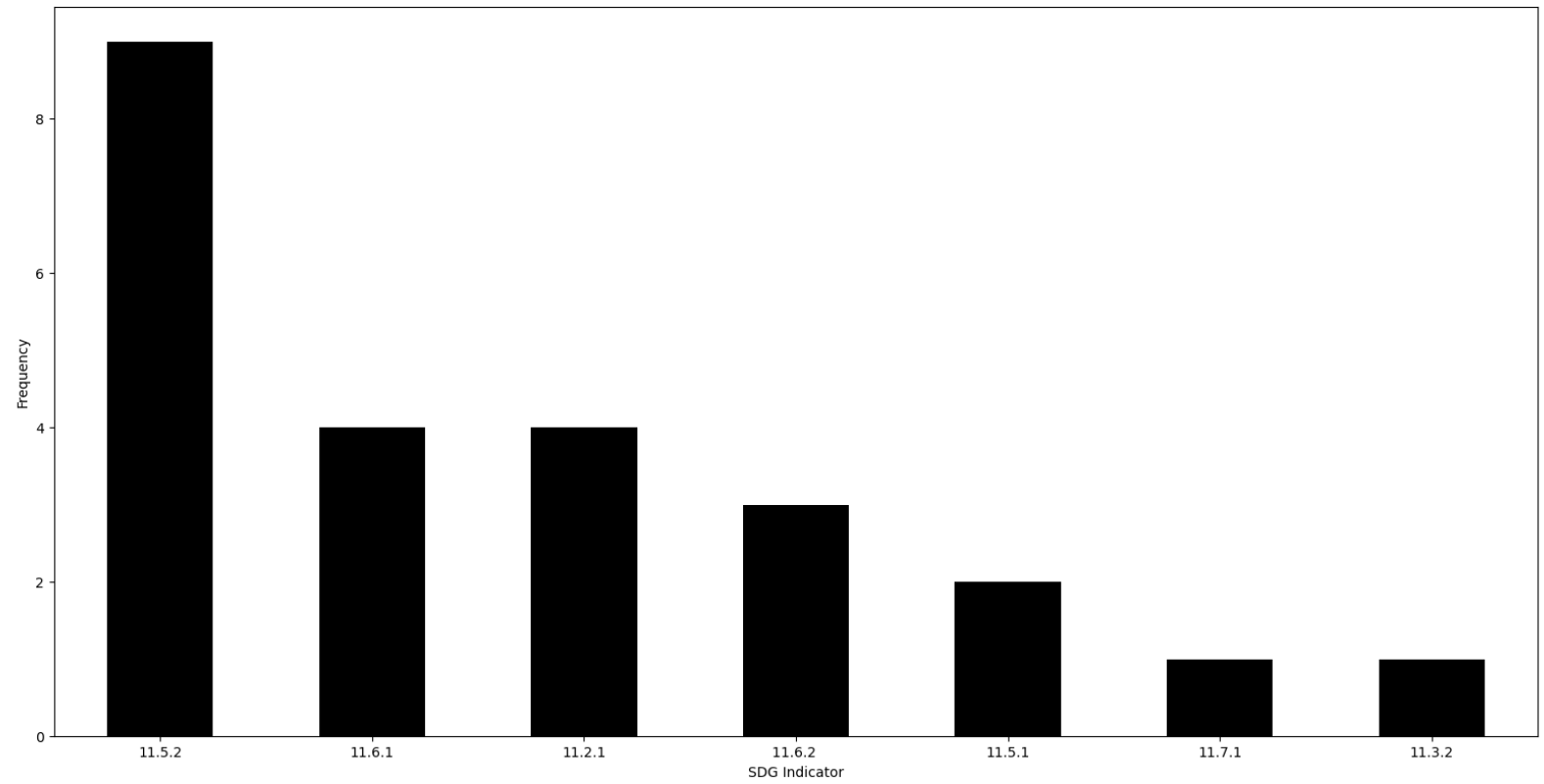

**Fig. 3** Projects in the dataset and their relevance to the SDG11 indicators.

[3] https://www.aiforsdgs.org/all-projects/aixai-area-information-x-artificial-intelligence.
[4] https://medium.com/astec/prometea-artificial-intelligence-in-the-judicial-system-of-argentina-4dfbde079c4

- *UNIST Heatwave Research for National Heat Wave Policy* (Target 11.3): the project uses artificial intelligence to investigate "detailed thermal characteristics of urban areas"[5]. None of Indicator 11.3.1 (ratio of land consumption rate to population growth rate) and Indicator 11.3.2 (proportion of cities with a direct participation structure of civil society in urban planning) would have reflected the value of the contribution. This seems to point at the fact that there are more aspects to sustainable urban planning than land consumption/population growth rate, and citizen participation alone;
- *geoland* (Target 3): the project provides several geoinformation services for land monitoring[6]. Same argument as just above regarding the aspects of sustainable urban planning;
- *CAMELS* (Target 3): the project focused on monitoring terrestrial carbon sink and its causes[7]. The argument above regarding the aspects of sustainable urban planning also applies.

## 4.6 Contribution to Citizen-centric challenges

Three projects had contributions relevant to the citizen-centric challenges. *Breeze* provides services in the area of air quality sensors, air quality data and air quality analytics. It offers citizens the opportunity to participate by becoming sensor hosts. This is an example of a measure to facilitate deeper citizen participation. The *MEDACTION 4* project contributed a Public

[5] https://www.aiforsdgs.org/all-projects/unist-heatwave-research-national-heat-wave-policy.
[6] https://cordis.europa.eu/project/id/502871/results
[7] https://cordis.europa.eu/article/id/85263-estimating-europes-carbon-dioxide-fluxes.

Participation Geographical Information System (PPGIS) featuring neural network components. Participatory stakeholder workshops were also organized to engage the public with the driving forces and effects of land degradation and desertification. This too is an example of a measure to promote deeper citizen participation. Finally, the *DAYWATER* project produced the Hydropolis app, which has different types of users (i.e., guest, user, manager, administrator) and adapts the level of information provided according to their background[8]. This could be seen as a primitive form of adaptivity/personalization. In general, the number of projects in the dataset which can be said to connect to the citizen-centric challenges is relatively low (10%).

# 5 Discussion

## 5.1 Key takeaways

As for the geographic distribution, there are some notable disparities, with Europe over-represented in this dataset and the rest of the world having fewer contributions. This may be a feature of the dataset or a true indication that other countries/continents are doing less regarding AI contributions to more sustainable cities. At this point, we attribute our observations to the fact that half of the data items came from the CORDIS database, which biases it automatically towards European cities. The value of this work is to have provided a snapshot that could be extended towards a more comprehensive picture of AI contributions to more sustainable cities worldwide.

Regarding the SDG targets and indicators, a noteworthy observation is that some targets did not appear at all in the sample. This is the case for Target 11.1 (safe and affordable housing), Target 11.4 (protection of the world's cultural and natural heritage), Target 11.a (strengthening economic, social and environmental links between urban, peri-urban and rural areas) and Target 11.c (support least developed countries in building sustainable and resilient buildings utilizing local materials). There are also two possible options: these areas have indeed received little attention so far (and hence it is worth exploring the opportunities of digitization to provide some added value), or the observation is a feature of the bias of the current dataset. Another takeaway from the dataset is that AI systems have indeed contributed to tackling issues of sustainable cities in several ways: waste management, air quality monitoring (and more broadly environmental monitoring), disaster response management, transportation management. Given that AI systems are on the rise, it can be expected that their number in the areas just mentioned and other areas of sustainable development will increase. Thus, it could be useful to explore ways of documenting best practices for AI implementation/deployment/use in these different areas. The issue is by no means trivial. There are, as the data has shown, different stakeholders with potentially conflicting interests (e.g., companies that may want to preserve what works as a competitive advantage, and research that wants to make knowledge available to all).

[8] See http://daywater.in2p3.fr/EN/guide/chapter6.php.

Finally, a key takeaway regarding the citizen-centric challenges is that many projects are still working *for* citizens (i.e., on their behalf) and not *with* them (i.e., actively involving them). A reason for this may be the fact that AI for social good is still in its infancy. For instance, several projects mentioned in Section 3 dealt with disaster mitigation. This is an endeavour for which the value of involving citizens has been documented in the past (e.g., Zook et al. (2010)). It may be conjectured that as the AI for social good initiatives mature, the involvement of citizens will become more pronounced.

## 5.2 Limitations

A general limitation of the current work is that it has been only descriptive and not explanatory (e.g., we can say little at the moment about why the state of affairs observed has been observed). We have also mentioned that the dataset is biased towards European cities. The method also has inherent limitations: 1) the assignment of locations to the projects a posteriori was subject to some reasonable assumptions but was still arbitrary to some extent: having those locations assigned a priori in a database would provide a more consistent picture of the geographical distributions; 2) the decision whether or not a project was AI-powered was made based on the keywords from the AI Glossary and the JRC: it may well be that some authors doing truly valuable AI work have not used these keywords in the descriptions used for the assessment (i.e., CORDIS and AI4SDG); 3) many projects from the CORDIS database were completed before 2015 when the SDG agenda was agreed upon and thus did not have the SDG goals in mind; and 4) the mapping of the projects to the SDG was done to the most relevant target: it would have been equally possible to list a project under several targets. Finally, we were deliberately interested in SDG11 and mapped the project to the targets and indicators related to SDG11. The contributions of some of the projects apply to more than SDG11, and extending our analysis might unveil interesting patterns about the synergies of SDGs (e.g., which contributions apply to which SDGs simultaneously and which SDGs share how many contributions more often).

## 5.3 Future work

As AI for social good is a new area, the interesting question is that of the evaluation of success. We have pondered this question at the beginning of the work but dropped it from the analysis because it was unclear from the documentation of most projects how the solutions were evaluated. In general, the task of empirically assessing the contributions has proven more challenging than expected because of the lack of homogeneous documentation. The present trend in literature also suggests the lack of reporting towards carbon emission and energy consumption, suggesting adverse impact to the sustainable development efforts (Henderson et al., 2020). There is thus an opportunity for initiatives that 1) offer an ongoing call for AI4SDG projects; 2) provide a simple, structured template for AI systems' developers to document the value of their work. Such initiatives will be critical in assessing where we are and advancing the science of AI for social good.

## 6 Conclusion

In this chapter, we analyzed the contributions of AI systems to cities and illuminated areas of AI4SG that deserve more attention on the road towards more sustainable cities for SDG 11 and beyond. To help understand the current impacts of AI, the analysis presents the geographic distribution of the AI projects, their key beneficiaries, system type, the target, and indicators for which they are relevant that could be supportive in gaining knowledge about the influence of AI on suitable targets and indicators, type of technologies, their social impact and responsible stakeholder. We have learned that AI systems have indeed contributed to advancing sustainable cities in several ways (e.g., waste management, air quality monitoring, disaster response management, transportation management), but many projects are still working *for* citizens and not *with* them. This current snapshot of the impact of AI projects on SDG 11 has been limited by the quantity and the quality of the available data on existing AI projects. As we move towards more mature work on AI for social good, initiatives that promote consistent and high-quality documentation of AI projects will be vital for a deeper understanding of AI's impact on more/less sustainable and inclusive cities.

## Availability of data and materials

The list of projects presented is available at https://doi.org/10.6084/m9.figshare.17008366.

## Acknowledgements

Shivam Gupta gratefully acknowledges funding provided by the German Federal Ministry for Education and Research (BMBF) for the project "digitainable". Auriol Degbelo gratefully acknowledges funding from the European Social Fund and the Ministry of Economic Affairs, Innovation, Digitalization and Energy of the State of North Rhine-Westphalia through the SmartLandMaps 2.0 project (EFRE-0400389).

## Appendix A

### Definition of the SDG 11 Targets found in the dataset

In alphabetical order.

- 11.b substantially increase the number of cities and human settlements adopting and implementing integrated policies and plans towards inclusion, resource efficiency, mitigation and adaptation to climate change, resilience to disasters, and develop and implement... holistic disaster risk management at all levels;
- 11.2 provide access to safe, affordable, accessible and sustainable transport systems for all, improving road safety, notably by expanding public transport, with special attention to the needs of those in vulnerable situations, women, children, persons with disabilities and older persons;

- 11.3 enhance inclusive and sustainable urbanization and capacity for participatory, integrated and sustainable human settlement planning and management in all countries;
- 11.5 significantly reduce the number of deaths and the number of people affected and substantially decrease the direct economic losses relative to global gross domestic product caused by disasters, including water-related disasters, with a focus on protecting the poor and people in vulnerable situations;
- 11.6 reduce the adverse per capita environmental impact of cities, including by paying special attention to air quality and municipal and other waste management;
- 11.7 provide universal access to safe, inclusive and accessible, green and public spaces, in particular for women and children, older persons and persons with disabilities.

**Appendix B**

**Definition of the SDG 11 Indicators found in the dataset**

In alphabetical order.

- 11.2.1 Proportion of population that has convenient access to public transport, by sex, age and persons with disabilities;
- 11.3.2 Proportion of cities with a direct participation structure of civil society in urban planning and management that operate regularly and democratically;
- 11.5.1 Number of deaths, missing persons and directly affected persons attributed to disasters per 100,000 populations;

- 11.5.2 Direct economic loss in relation to global GDP, damage to critical infrastructure and number of disruptions to basic services, attributed to disasters;
- 11.6.1 Proportion of municipal solid waste collected and managed in controlled facilities out of total municipal waste generated, by cities;
- 11.6.2 Annual mean levels of fine particulate matter (e.g., PM2.5 and PM10) in cities (population weighted);
- 11.7.1 Average share of the built-up area of cities that is open space for public use for all, by sex, age and persons with disabilities.

**References**

Acuto, M. 2016. Give cities a seat at the top table. *Nature News 537*(7622): 611.

Allam, Z. and Z.A. Dhunny. 2019. On big data, artificial intelligence and smart cities. *Cities* 89: 80–91.

Allam, Z. and D.S. Jones 2020. On the coronavirus (covid-19) outbreak and the smart city network: universal data sharing standards coupled with artificial intelligence (ai) to benefit urban health monitoring and management. In *Healthcare*, Volume 8, pp. 46. Multidisciplinary Digital Publishing Institute.

Antweiler, C. 2019. Local knowledge theory and methods: an urban model from indonesia, *Investigating Local Knowledge*, 1–34. Routledge.

Aust, H.P. 2019. The shifting role of cities in the global climate change regime: From paris to pittsburgh and back? *Review of European, Comparative & International Environmental Law 28*(1): 57–66.

Axinte, L.F., A. Mehmood, T. Marsden, and D. Roep. 2019. Regenerative city-regions: a new conceptual framework. *Regional Studies, Regional Science 6*(1): 117–129.

Barlacchi, G., M. De Nadai, R. Larcher, A. Casella, C. Chitic, G. Torrisi, F. Antonelli, A. Vespignani, A. Pentland, and B. Lepri. 2015. A multisource dataset of urban life in the city of milan and the province of trentino. *Scientific data 2*(1): 1–15.

Barns, S. 2019. *Platform urbanism: negotiating platform ecosystems in connected cities*. Springer.

Batty, M. 2009. Cities as complex systems: Scaling, interaction, networks, dynamics and urban morphologies.

Bibri, S.E. 2021. Data-driven smart sustainable cities of the future: An evidence synthesis approach to a comprehensive state-of-the-art literature review. *Sustainable Futures*: 100047.

Chase, A.C. 2020. Ethics of ai: Perpetuating racial inequalities in healthcare delivery and patient outcomes. *Voices in Bioethics* 6.

Chipofya, M., M. Karamesouti, C. Schultz, and A. Schwering. 2020. Local domain models for land tenure documentation and their interpretation into the LADM. *Land Use Policy* 99: 105005. https://doi.org/https://doi.org/10.1016/j.landusepol.2020.105005 .

Cowls, J., A. Tsamados, M. Taddeo, and L. Floridi. 2021a, feb. A definition, benchmark and database of AI for social good initiatives. *Nature Machine Intelligence 3*(2): 111–115. https://doi.org/10.1038/s42256-021-00296-0 .

Cowls, J., A. Tsamados, M. Taddeo, and L. Floridi. 2021b. A definition, benchmark and database of ai for social good initiatives. *Nature Machine Intelligence 3*(2): 111–115.

Croese, S., C. Green, and G. Morgan. 2020. Localizing the sustainable development goals through the lens of urban resilience: Lessons and learnings from 100 resilient cities and cape town. *Sustainability 12*(2): 550.

Dai, W. 2019. Quantum-computing with ai & blockchain: modelling, fault tolerance and capacity scheduling. *Mathematical and Computer Modelling of Dynamical Systems 25*(6): 523–559.

Degbelo, A., C. Granell, S. Trilles, D. Bhattacharya, S. Casteleyn, and C. Kray. 2016. Opening up smart cities: citizen-centric challenges and opportunities from GIScience. *ISPRS International Journal of Geo-Information 5*(2): 16. https://doi.org/10.3390/ijgi5020016.

Degbelo, A., C. Stöcker, K. Kundert, and M. Chipofya 2021. SmartLandMaps - From customary tenure to land information systems. In *FIG e-Working Week 2021 – Challenges in a new Reality*.

Dinh, T.N. and M.T. Thai. 2018. Ai and blockchain: A disruptive integration. *Computer 51*(9): 48–53.

Ferri, M., U. Wehn, L. See, M. Monego, and S. Fritz. 2020. The value of citizen science for flood risk reduction: cost–benefit analysis of a citizen observatory in the brenta-bacchiglione catchment. *Hydrology and Earth System Sciences 24*(12): 5781–5798.

Firouzi, F., B. Farahani, and A. Marinˇsek. 2021. The convergence and interplay of edge, fog, and cloud in the ai-driven internet of things (iot). *Information Systems*: 101840.

Fitjar, R.D. and A. Rodr´ıguez-Pose. 2020. Where cities fail to triumph: The impact of urban location and local collaboration on innovation in norway. *Journal of Regional Science 60*(1): 5–32.

Fraisl, D., J. Campbell, L. See, U. Wehn, J. Wardlaw, M. Gold, I. Moorthy, R. Arias, J. Piera, J.L. Oliver, et al. 2020. Mapping citizen science contributions to the un sustainable development goals. *Sustainability Science 15*(6): 1735–1751.

Fritz, S., L. See, T. Carlson, M.M. Haklay, J.L. Oliver, D. Fraisl, R. Mondardini, M. Brocklehurst, L.A. Shanley, S. Schade, et al. 2019. Citizen science and the united nations sustainable development goals. *Nature Sustainability 2*(10): 922–930.

Galaz, V., M.A. Centeno, P.W. Callahan, A. Causevic, T. Patterson, I. Brass, S. Baum, D. Farber, J. Fischer, D. Garcia, et al. 2021. Artificial intelligence, systemic risks, and sustainability. *Technology in Society* 67: 101741.

Guan, T., K. Meng, W. Liu, and L. Xue. 2019. Public attitudes toward sustainable development goals: Evidence from five chinese cities. *Sustainability 11*(20): 5793.

Gupta, S., A. Hamzin, and A. Degbelo. 2018, oct. A low-cost open hardware system for collecting traffic data using Wi-Fi signal strength. *Sensors 18*(11): 3623. https://doi.org/10.3390/s18113623 .

Gupta, S., S.D. Langhans, S. Domisch, F. Fuso-Nerini, A. Fell¨ander, M. Battaglini, M. Tegmark, and R. Vinuesa. 2021. Assessing whether artificial intelligence is an enabler or an inhibitor of sustainability at indicator level. *Transportation Engineering* 4: 100064.

Gupta, S., E. Pebesma, J. Mateu, and A. Degbelo. 2018. Air quality monitoring network design optimisation for robust land use regression models. *Sustainability 10*(5): 1442. https://doi.org/10.3390/su10051442.

Henderson, P., J. Hu, J. Romoff, E. Brunskill, D. Jurafsky, and J. Pineau. 2020. Towards the systematic reporting of the energy and carbon footprints of machine learning. *Journal of Machine Learning Research 21*(248): 1–43.

Hilbert, M. 2016. Big data for development: A review of promises and challenges. *Development Policy Review 34*(1): 135–174.

Hutson, M. 2017, jul. AI Glossary: Artificial intelligence, in so many words. *Science 357*(6346): 19–19. https://doi.org/10.1126/science.357.6346.19 .

Ismagilova, E., L. Hughes, Y.K. Dwivedi, and K.R. Raman. 2019. Smart cities: Advances in research—an information systems perspective. *International Journal of Information Management* 47: 88–100.

Israilidis, J., K. Odusanya, and M.U. Mazhar. 2021. Exploring knowledge management perspectives in smart city research: A review and future research agenda. *International Journal of Information Management* 56: 101989.

Kirwan, C.G. and F. Zhiyong. 2020. *Smart Cities and Artificial Intelligence: Convergent Systems for Planning, Design, and Operations*. Elsevier.

Kuffer, M., D.R. Thomson, G. Boo, R. Mahabir, T. Grippa, S. Vanhuysse, R. Engstrom, R. Ndugwa, J. Makau, E. Darin, et al. 2020. The role of earth observation in an integrated deprived area mapping "system" for low-tomiddle income countries. *Remote sensing 12*(6): 982.

Kuffer, M., J. Wang, D.R. Thomson, S. Georganos, A. Abascal, M. Owusu, and S. Vanhuysse. 2021. Spatial information gaps on deprived urban areas (slums) in low-and-middle-income-countries: A user-centered approach. *Urban Science 5*(4): 72.

Li, L., X. Xia, B. Chen, and L. Sun. 2018. Public participation in achieving sustainable development goals in china: Evidence from the practice of air pollution control. *Journal of cleaner production* 201: 499–506.

Magni, G. 2017. Indigenous knowledge and implications for the sustainable development agenda. *European Journal of Education 52*(4): 437–447.

Majumdar, S., M.M. Subhani, B. Roullier, A. Anjum, and R. Zhu. 2021. Congestion prediction for smart sustainable cities using iot and machine learning approaches. *Sustainable Cities and Society* 64: 102500.

Makondo, C.C. and D.S. Thomas. 2018. Climate change adaptation: Linking indigenous knowledge with western science for effective adaptation. *Environmental science & policy* 88: 83–91.

Martens, J. 2019. Revisiting the hardware of sustainable development. *Reshaping*: 11.

Micheletti, M., D. Stolle, and D. Berlin. 2014. Sustainable citizenship: The role of citizens and consumers as agents of the environmental state. *State and environment: The comparative study of environmental governance*: 203–236.

Munsaka, E. and E. Dube. 2018. The contribution of indigenous knowledge to disaster risk reduction activities in zimbabwe: A big call to practitioners. *Jàmbá: Journal of Disaster Risk Studies 10*(1): 1–8.

Newman, G., T. Shi, Z. Yao, D. Li, G. Sansom, K. Kirsch, G. Casillas, and J. Horney. 2020. Citizen science-informed community master planning: Land use and built environment changes to increase flood resilience and decrease contaminant exposure. *International journal of environmental research and public health 17*(2): 486.

Nilsson, M., D. Griggs, and M. Visbeck. 2016. Policy: map the interactions between sustainable development goals. *Nature News 534*(7607): 320.

Palomares, I., E. Martínez-Cámara, R. Montes, P. Garc´ıa-Moral, M. Chiachio, J. Chiachio, S. Alonso, F.J. Melero, D. Molina, B. Fern´andez, et al. 2021. A panoramic view and swot analysis of artificial intelligence for achieving the sustainable development goals by 2030: progress and prospects. *Applied Intelligence*: 1–31.

Pekmezovic, A. 2019. The un and goal setting: From the mdgs to the sdgs. *Sustainable Development Goals: Harnessing Business to Achieve the SDGs through Finance, Technology, and Law Reform*: 17–35.

Rabah, K. 2018. Convergence of ai, iot, big data and blockchain: a review. *The lake institute Journal 1*(1): 1–18.

Rajan, K. and A. Saffiotti. 2017. Towards a science of integrated ai and robotics.

Reddick, C.G., R. Enriquez, R.J. Harris, and B. Sharma. 2020. Determinants of broadband access and affordability: An analysis of a community survey on the digital divide. *Cities* 106: 102904.

Rogers, B., G. Dunn, K. Hammer, W. Novalia, F. de Haan, L. Brown, R. Brown, S. Lloyd, C. Urich, T. Wong, et al. 2020. Water sensitive cities index:

A diagnostic tool to assess water sensitivity and guide management actions. *Water research* 186: 116411.

Rubio-Mozos, E., F.E. García-Muiña, and L. Fuentes-Moraleda. 2019. Rethinking 21st-century businesses: An approach to fourth sector smes in their transition to a sustainable model committed to sdgs. *Sustainability 11*(20): 5569.

Salehi, M., L.I. Rusu, T.M. Lynar, and A. Phan 2016. Dynamic and robust wildfire risk prediction system: An unsupervised approach. In B. Krishnapuram, M. Shah, A. J. Smola, C. C. Aggarwal, D. Shen, and R. Rastogi (Eds.), *Proceedings of the 22nd ACM SIGKDD International Conference on Knowledge Discovery and Data Mining (KDD 2016)*, San Francisco, California, USA, pp. 245–254. ACM.

Samoili, S., M. López-Cobo, E. Gómez, G. De Prato, F. Mart´ınez-Plumed, and B. Delipetrev. 2020. AI Watch defining Artificial Intelligence. *Publications Office of the European Union.*

Saner, R., L. Yiu, and M. Nguyen. 2020. Monitoring the sdgs: digital and social technologies to ensure citizen participation, inclusiveness and transparency. *Development Policy Review 38*(4): 483–500.

Șerban, A.C. and M.D. Lytras. 2020. Artificial intelligence for smart renewable energy sector in europe—smart energy infrastructures for next generation smart cities. *IEEE Access* 8: 77364–77377.

Shahid, N., M.A. Shah, A. Khan, C. Maple, and G. Jeon. 2021. Towards greener smart cities and road traffic forecasting using air pollution data. *Sustainable Cities and Society*: 103062.

Sharda, S., M. Singh, and K. Sharma. 2021. Demand side management through load shifting in iot based hems: Overview, challenges and opportunities. *Sustainable Cities and Society* 65: 102517.

Solecki, W., C. Rosenzweig, S. Dhakal, D. Roberts, A.S. Barau, S. Schultz, and D. Urge-Vorsatz. 2018. City transformations in a 1.5 c warmer world. ¨ *Nature Climate Change 8*(3): 177–181.

Sougkakis, V., K. Lymperopoulos, N. Nikolopoulos, N. Margaritis, P. Giourka, and K. Angelakoglou. 2020. An investigation on the feasibility of near-zero and positive energy communities in the greek context. *Smart Cities 3*(2): 362–384.

Taddeo, M., A. Tsamados, J. Cowls, and L. Floridi. 2021. Artificial intelligence and the climate emergency: Opportunities, challenges, and recommendations. *One Earth 4*(6): 776–779.

Tajunisa, M., L. Sadath, and R.S. Nair. 2021. Nanotechnology and artificial intelligence for precision medicine in oncology, *Artificial Intelligence*, 103–122. CRC Press.

Thinyane, M. 2018. Engaging citizens for sustainable development: a data perspective.

Tomašev, N., J. Cornebise, F. Hutter, S. Mohamed, A. Picciariello, B. Connelly, D.C. Belgrave, D. Ezer, F.C. van der Haert, F. Mugisha, et al. 2020. Ai for social good: unlocking the opportunity for positive impact. *Nature Communications 11*(1): 1–6.

UNDESA Statistics Division. 2021. SDG Indicators Metadata repository. https://unstats.un.org/sdgs/metadata/?Text=&Goal=11&Target=. Accessed: 2021-09-30.

United Nations, 2015. Sustainable development goals: 17 goals to transform our world. *United Nations, [Online]. Available: https://www. un. org/sustainabledevelopment/energy/. [Accessed 04 June 2018].*

United Nations Department of Economic and Social Affairs (UNDESA). 2015. Transforming our world: The 2030 agenda for sustainable development.

van Wynsberghe, A. 2021. Sustainable ai: Ai for sustainability and the sustainability of ai. *AI and Ethics*: 1–6.

Villagra, A., E. Alba, and G. Luque. 2020. A better understanding on traffic light scheduling: New cellular gas and new in-depth analysis of solutions. *Journal of Computational Science* 41: 101085.

Vinuesa, R., H. Azizpour, I. Leite, M. Balaam, V. Dignum, S. Domisch, A. Fell¨ander, S.D. Langhans, M. Tegmark, and F.F. Nerini. 2020. The role of artificial intelligence in achieving the sustainable development goals. *Nature communications 11*(1): 1–10.

Walker, J., A. Pekmezovic, and G. Walker. 2019. *Sustainable development goals: harnessing business to achieve the SDGs through finance, technology and law reform*. John Wiley & Sons.

Yatoo, S.A., P. Sahu, M.H. Kalubarme, and B.B. Kansara. 2020. Monitoring land use changes and its future prospects using cellular automata simulation and artificial neural network for ahmedabad city, india. *GeoJournal*: 1–22.

Yigitcanlar, T., M. Kamruzzaman, M. Foth, J. Sabatini-Marques, E. da Costa, and G. Ioppolo. 2019. Can cities become smart without being sustainable? a systematic review of the literature. *Sustainable cities and society* 45: 348–365.

Zheng, C., J. Yuan, L. Zhu, Y. Zhang, and Q. Shao. 2020. From digital to sustainable: A scientometric review of smart city literature between 1990 and 2019. *Journal of Cleaner Production* 258: 120689.

Zook, M., M. Graham, T. Shelton, and S. Gorman. 2010. Volunteered geographic information and crowdsourcing disaster relief: a case study of the Haitian earthquake. *World Medical Health Policy 2*(2): 2. https://doi.org/10.2202/1948-4682.1069.